\documentclass[letterpaper]{article}
\usepackage{iccc}
\usepackage{natbib}
\usepackage[most]{tcolorbox}
\usepackage{listings}
\usepackage{algorithm}
\usepackage{algpseudocode}
\usepackage{booktabs}
\usepackage{graphicx}
\usepackage{subcaption}
\usepackage{arydshln}

\usepackage{times}
\usepackage{helvet}
\usepackage{courier}
\title{Evolving to the Aesthetics of a Vision-Language Model}
\author{Stephen James Krol \\
        SensiLab, Monash University \\
        Melbourne, Australia \\
        stephen.krol@monash.edu
        \And 
        Jon McCormack \\ 
        SensiLab, Monash University \\
        Melbourne, Australia \\
        jon.mccormack@monash.edu
    }
\begin{document} 
\maketitle
\begin{abstract}
\begin{quote}
Evolutionary systems have demonstrated remarkable results in creative domains, with recent applications in generative typography, design, and music. However, an open problem remains in designing fitness functions that effectively capture the desired aesthetics of abstract outputs. In this work, we explore two methods for evaluating the aesthetics of a population using Vision-Language Models (VLMs). The first method uses CLIP-IQA to predict an aesthetic score for each design. The second method instead pits candidates against each other, with winners determined by a VLM using a custom prompt specified by the user. The outcomes of these pairwise comparisons are then used to estimate a population ranking via the Glicko rating system. We present these methods in the context of a case study using a custom generative system and compare the resulting rankings with an artist’s aesthetic ranking and those produced by other aesthetic evaluation techniques. Additionally, we document the artist’s experience using these approaches to evolve designs, critically analysing the strengths and weaknesses of both methods.
\end{quote}
\end{abstract}

\section{Introduction}

The development of algorithms for Image Aesthetic Quality Assessment (IAQA) has been widely studied in both Artificial Intelligence (AI) and Computational Creativity (CC) \citep{wang2023exploring,xiong2024image,lopes2023towards,goree2021does}. Within this broader context, Evolutionary Computing (EC) has addressed the problem by designing aesthetic fitness functions to guide the evolution of outputs from parametric models \citep{Mccormack2023,Tatsu02013,machado2013fitness,rebelo2020evolutionary}. Approaches include the development of handcrafted aesthetic rules \citep{lopes2023towards}, the use of proxy measures to estimate aesthetic quality \citep{mccormack2022complexity}, and, more recently, the application of large pre-trained Deep Learning (DL) models to score artefacts \citep{Gomide2025,wang2023exploring}.

Although the recent rise of diffusion models \citep{ho2020denoising,rombach2022high} has overshadowed more traditional generative systems, parametric models remain relevant in both artistic practice \citep{Latham2025,lomas2016species} and design \citep{snooks2021behavioral}, making their continued study worthwhile in the current technological climate. In this context, automatic IAQA facilitates the identification of high-quality designs within complex systems whose search spaces are often too large to explore effectively by hand, thereby improving the overall usability of these systems.

In this paper, we investigate two IAQA paradigms for EC based on pre-trained (DL) models. The first is a point-wise evaluation approach using the state-of-the-art (SOTA) CLIP-IQA method \citep{wang2023exploring}, which generates an aesthetic score from custom antonym prompt pairings. The second employs a pre-trained \textit{Qwen3-VL-8B} Vision-Language Model (VLM) to perform pairwise comparisons between images using a custom prompt; this process is repeated across the population, and the results are aggregated into a global ranking using the Glicko rating system \citep{glickman1995glicko}.
While prior work has demonstrated that pre-trained DL models can be effective in evaluating the aesthetics of representational images \citep{wang2023exploring,Gomide2025}, comparatively little research has examined their effectiveness for non-representational artefacts \citep{lomas2016species}. These approaches are therefore evaluated against a custom-built benchmark to assess their performance in ranking non-representational images.

Finally, both methods are applied to a real parametric system in an artist-centred study, in which an artist used them to evolve designs from a custom Harmonograph-inspired software system, providing practical insight into their value within active creative practice. Results indicate that while both the point-wise and pairwise approaches achieved comparable performance on the benchmark, the pairwise VLM method afforded the artist greater control than the point-wise CLIP-IQA approach, albeit at a substantially higher computational cost. Hence, the contributions of this work are as follows:

\begin{itemize}
    \item A comparison of two IAQA paradigms in evaluating the aesthetics of non-representational imagery. 
    \item An artist-centred study on the use of pre-trained VLM models to evolve the design space of a parametric system.
\end{itemize}

The code for this project is also available online\footnote{https://github.com/SensiLab/aesthetic-evolution}.

\section{Related Work}

\subsection{Evaluating the Aesthetics of Images}

The field of Image Quality Assessment (IQA) has a multitude of methods designed to evaluate image features such as noise level and image distortion \citep{Zhou2004}, with several methods demonstrating strong performance in evaluating general image quality \citep{Wu2024,chen2024topiq}. Additionally, the field has explored methods for IAQA, which aims to estimate more subjective image qualities such as aesthetics \citep{wang2023exploring} or perceived emotion \citep{Parthasarathy2017}. For example, \citet{datta2006studying} analysed photographs to develop various visual features to train Machine Learning (ML) models on. Other techniques instead train or leverage large pre-trained deep learning (DL) models to build aesthetic scores \citep{xiong2024image}. \citet{wang2023exploring} are a notable example, introducing their method CLIP-IQA which utilises a pre-trained CLIP model \citep{radford2021learning} to score various aesthetic attributes of an image using antonym prompt pairings. Other techniques, such as that presented in \citet{xiong2024image}, perform additional training using various benchmark datasets \citep{murray2012ava, fang2020perceptual}, to estimate aesthetic scores from human annotated data. However, while these DL methods have demonstrated SOTA performance on various benchmarks, their applications have mostly been on representational images, with their performance on more abstract imagery being untested. Furthermore, the prediction strategy of condensing aesthetics to a score, or set of scores, is at odds with various theories on how humans measure \textit{feel} \citep{rokeach1973nature,kahneman2013perspective}. \citet{yannakakis2018ordinal} argue for this, providing a multi-disciplinary perspective on why emotions are \textit{ordinal} and highlight the value of preference learning \citep{burges2005learning,Yannakakis2009} as a more suitable method for predicting \textit{feel}. Furthermore, recent work \citep{chen2024mllm} has demonstrated that multi-modal VLMs align with human judgements in pairwise comparisons on vision-language benchmarks providing motivation to explore their use in aesthetic ranking. Our work builds on prior research by evaluating and comparing point-wise and pairwise approaches to aesthetic ranking using pre-trained DL models, and by applying both techniques within an artist-centred study.

\subsection{Aesthetics and Computational Creativity}

Within CC, various works have investigated and commented on the concept of aesthetics and its importance in creativity. This has moved beyond just evaluating the quality of outputs to formalising and developing systems that are capable of defining their own aesthetic. \citet{colton2008creativity} argues that evaluating a system solely by its generated artefacts is inadequate, noting that knowledge of the process impacted one's perception of the artefact's aesthetic \citep{colton2012painting}. \citet{guckelsberger2017addressing} extend this, highlighting that the to improve perceived creativity, systems must also be able to explain \textit{why} they acted in a particular manner and thus should have internal goals and motivations that are not directly tied to the system's designer. \citet{bodily2018explainability} further argue that "creativity is an inherently social construct" which requires those participating to communicate their intention and describe their developed aesthetic. 

Compared to these works, this paper does not design a system capable of developing its own aesthetic, but instead attempts to leverage learned aesthetics from pre-trained VLMs. This positions the system as a tool that can assist users in exploring a generative system's search space.

\subsection{Evolving to Aesthetics}

Work on evolving for aesthetic outcomes takes many forms, but a common strategy has been to encode aesthetics explicitly in the fitness function. Earlier systems often relied on hand-crafted measures or rules to operationalize aesthetic criteria \citep{Tatsu02013,machado2013fitness}. For example, \citet{rebelo2020evolutionary} evolve a custom typographic system using three fitness functions targeting legibility, aesthetic quality, and semantic coherence, with their aesthetic component computed as the arithmetic mean of five bespoke attributes designed for document-level aesthetic analysis.

Other work has instead searched for automatic proxies that could be used to estimate aesthetics. For example, \citet{mccormack2022complexity} tested multiple complexity measures on evolutionary art datasets and found that correlations with aesthetic judgement vary by dataset and by measure, i.e., there is no universally ``best'' complexity metric for aesthetics; instead, usefulness depends on the system and the context of judgement. Here the authors framed the problem as less about discovering a universal aesthetic fitness and more about choosing (or learning) proxies that are locally valid for a given artist, medium, or generative process. 

With modern VLMs, this idea can be extended further: aesthetic fitness functions can be instantiated dynamically through prompting \citep{wang2023exploring}, enabling artists to guide evolutionary search with natural-language descriptions of their preferences. In their art installation, \citet{Latham2025} demonstrated the potential for this by ``fully [handing] over aesthetic and content decisions" to Google's Gemini and highlighting how its use as a ``selector" resulted in the evolution of various forms using a custom generative system. In our work, we extend this paradigm by studying an artist’s use of prompt-conditioned aesthetic fitness to generate non-representational visual art, moving beyond the exhibition context to evaluate the approach’s practical effectiveness.

\section{Ranking Systems}

\subsection{Point-Wise Ranking via CLIP-IQA}

CLIP-IQA \citep{wang2023exploring} is method for IQA that utilises a pre-trained CLIP model \citep{radford2021learning} --- a VLM trained to align images and text in a shared embedding space ---  to score images on various dimensions of \textit{feel}. To achieve this, simple antonym prompt pairings are designed to capture the different extremes of a concept. For example, \textit{Complex} \& \textit{Simple}, \textit{Good Picture} \& \textit{Bad Picture} and \textit{Happy} \& \textit{Sad}. Both prompts and their respective image are then passed through the CLIP model to retrieve $\textbf{x} \in R^c, \textbf{t}_1 \in R^c$ and $t_2 \in R^c$ which are the vector features of the image, positive prompt and negative prompt. The cosine similarity between each prompt vector and the image vector is then calculated as below:

\begin{equation}
    s_i = \frac{x \cdot t_i}{|x||t_i|}, i \in \{1, 2\}
\end{equation}

As CLIP is trained so that similar concepts result in vectors with similar direction, the cosine similarity function above measures the semantic relationship between the prompts and the image. These scores are then inputted into the Softmax function to produce the final score $\bar{s} \in [0, 1]$:

\begin{equation}
    \bar{s} = \frac{e^{s_1}}{e^{s_1} + e^{s_2}}
\end{equation}

This softmax technique is used to mitigate the ambiguity issue associated with using CLIP models for scoring. As argued by the authors, a \textit{rich} image could be one that depicts wealth or is rich in content. This antonym approach produces a relative score and as demonstrated in \citet{wang2023exploring}, results in significantly better correlation with human perception compared to single prompt setups. 

\subsection{Pairwise Judging via a VLM}

To rank designs by custom aesthetic criteria we utilised an off-the-shelf VLM \textit{Qwen3-VL-8B-Instruct} to act as a judge. The choice of VLM model was arbitrary and in theory any VLM that can accept multiple image inputs and text could be used. The VLM is provided three inputs: image 1 ($i_1$), image 2 ($i_2$) and a custom text prompt defining the aesthetic criteria to be evaluated ($P$). The user is given freedom in how they define the prompt but must ensure that the model is instructed to eventually decide which image better suits the desired aesthetic criteria. An example of a prompt is shown in Figure~1.

\begin{figure}
\begin{tcolorbox}[
    title=Example Prompt,
    colback=gray!5,
    colframe=gray!60,
    boxrule=0.5pt,
    arc=2pt
]

    You will be given two images depicting harmonograph drawings, you must choose which one is more aesthetically pleasing in representing harmonic patterns.

    IMPORTANT RULES (must be followed):
    \begin{itemize}
        \item Images that are mostly dark blobs or solid dark regions MUST be ranked lower.
        \item Visible line structure and repeating patterns are REQUIRED for a high score.
        \item Messy noise or amorphous shapes should be ranked lower.
        \item Winning designs must possess the best INTRICATE and CLEAR harmonic patterns.
    \end{itemize}
    
    Task:
    \newline
    Choose which image is more aesthetically pleasing according to the rules above. If you cannot make a decision return a draw value of `3'.
    Output a one sentence description of each image, a single sentence regarding your reasoning and finally a single digit corresponding to which image is better ONLY `1' or `2' or `3' for draw.

\end{tcolorbox}
\label{fig:prompt-example}
\caption{An example prompt used by the VLM}
\end{figure}

While any criteria can be used, its worth discussing two features of the above prompt that helped improve results: (1) Forcing the model to first describe the images and rationalise its decision before choosing an outcome resulted in better performance as demonstrated in Table \ref{tab:benchmark} and, (2) providing the option for a draw prevented the model from entering into continuous reasoning loops when there was no clear winning image.

A limitation of ranking designs using pairwise comparisons is the compute complexity which grows $O(N^2)$. This is particularly evident when working with VLMs which require specialised GPU hardware to enable fast inference. As briefly discussed in the next section, various engineering decisions were made to improve the inference speed of the VLM. However, previous work has demonstrated that global ranking can be estimated by sampling possible pairwise comparisons \citep{Baltrusaitis2017} suggesting that not all pairs need to be evaluated to estimate a ``good enough'' ranking.

On this basis, we implement the Glicko system to calculate the ranking of designs based on the outcomes of $n$ pairwise comparisons. The Glicko system \citep{glickman1995glicko} is a rating method for competitive games that estimates a player’s skill using both a rating and a rating deviation (RD), which reflects the uncertainty in that estimate. After each set of matches, ratings are updated based on game outcomes and the opponent’s rating and RD, allowing the system to adjust more quickly for players with higher uncertainty. The Glicko system has been use in similar contexts \citep{mccormack2022complexity} making it a suitable starting point for this task.

\section{Case Study: Generative Line Drawing}

To study the use of these ranking techniques in evaluating non-representational imagery, we present an artist-centred study that applies both methods to rank a set of images generated by a bespoke generative software system developed by the artist. We demonstrate the capabilities of the two techniques through quantitative benchmarking against the artist’s preferences, along with an analysis of the artist’s experience using the system to create new designs with a generative algorithm (GA).

\begin{figure*}
    \centering
    \includegraphics[width=\linewidth]{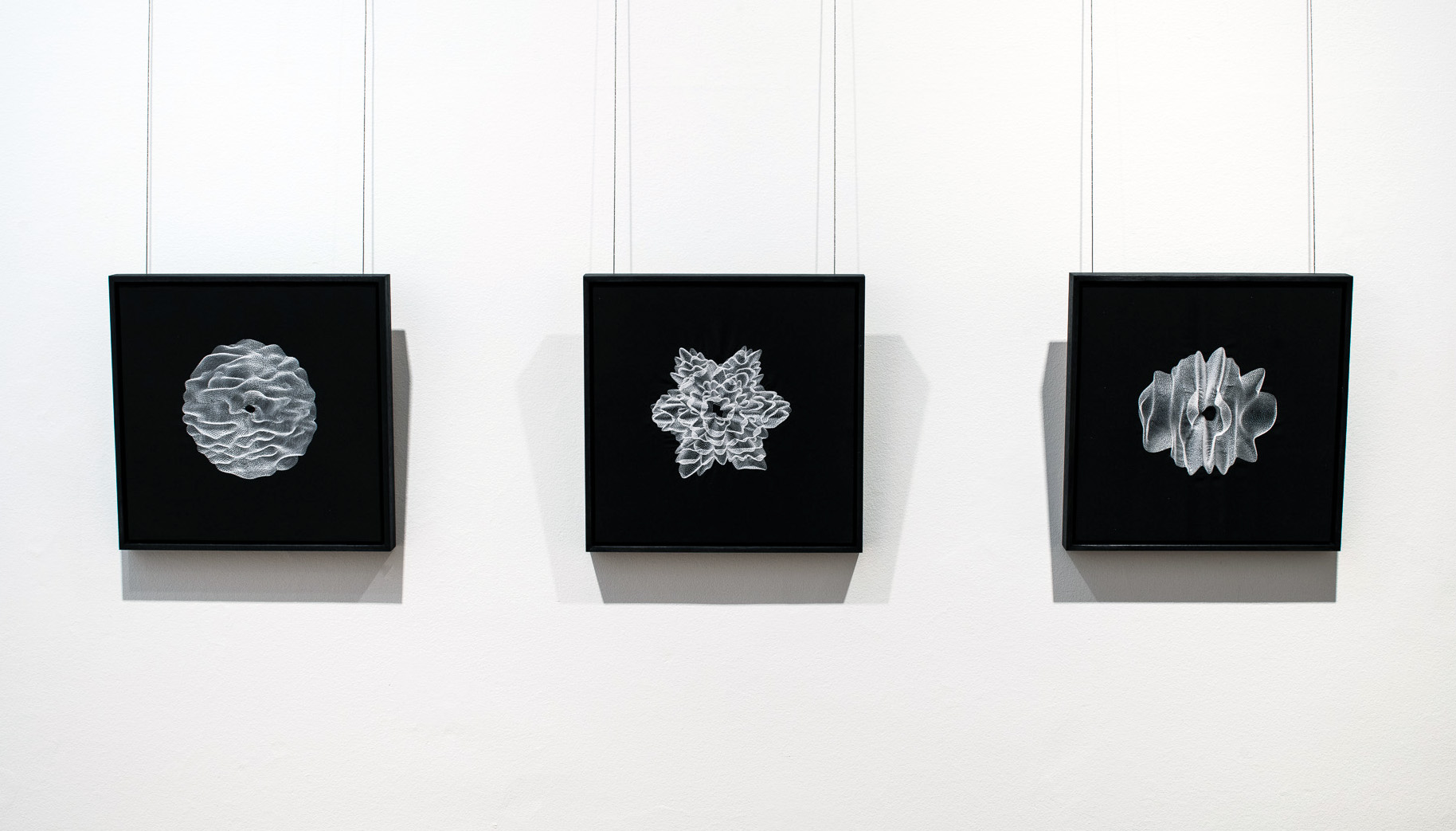}
    \caption{Example gallery exhibition of outputs from the generative line drawing system used in this paper. The drawings created by the software are converted to stitched cotton embroideries, machine embroided onto cotton fabric.}
    \label{fig:exhibition}
\end{figure*}

The artist's generative system, detailed in the next section, was selected for this study for several reasons. Firstly, the system has ``ecological validity'' \citep{Brunswik1956,schmucklerWhatEcologicalValidity2001,McCormackPBR2021} in that it has already been extensively used by a professional artist who has exhibited and sold works made by the system in commercial galleries (Figure~\ref{fig:exhibition}). Secondly, the system is relatively manageable, with only 17 continuous parameters, whose effects can be explored interactively by the software in real-time. Nonetheless, the search space -- a 17-dimension vector space -- is complex enough to make it impossible for the human artist to exhaustively search. The drawing system outputs line drawings in SVG format, allowing for 2D rasterization at any resolution, enabling evaluation via image-focused deep learning models. Lastly, the system generates abstract rather than figurative or representational imagery, making it a challenge for many deep-learning systems which are often trained on largely figurative imagery \citep{willisonExploringTrainingData}, and which are elusive when trying to describe using descriptive language, a requirement of prompt-based models \citep{mccormackWritingPromptsReally2023}.

\subsection{Custom Generative Art System}

\begin{figure}
    \centering
    \includegraphics[width=\linewidth]{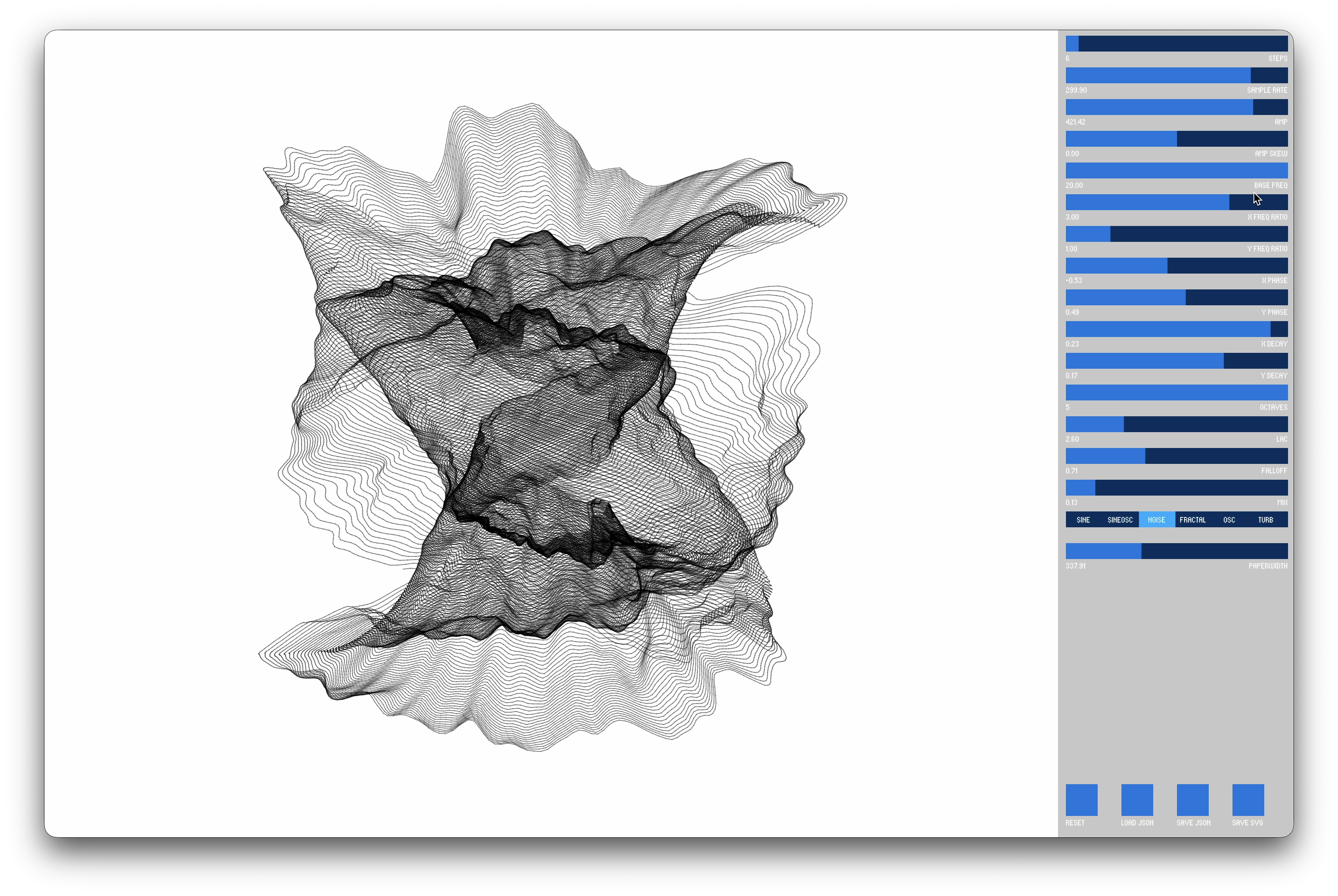}
    \caption{The artist-developed generative drawing system used in this case study. The figure shows current drawing (left) the interface and parameters which can be adjusted in real-time (right). }
    \label{fig:harmonograph_ide}
\end{figure}

Our custom generative drawing software simulates an oscillating line drawing agent, whose movements are controlled by a series of variable oscillators. Drawing inspiration from the \textit{harmonograph} -- a 19th century mechanical drawing device based on multiple swinging pendulums -- the software version uses a combination of periodic functions and aperiodic stationary noise \citep{Perlin2002}. This combination of functions allows a much richer variety of drawings over any mechanical harmonograph. To further increase the visual possibilities, the agent's drawing space can be non-linearly distorted, giving rise to additional creative possibilities. The drawing software was implemented in the Processing environment \citep{fryProcessingProgrammingHandbook2014} and uses an interactive interface with the 17 adjustable parameters controlling the agent's behaviour and manipulated using controls on the right (Figure \ref{fig:harmonograph_ide}). 

\subsubsection{Genetic Algorithm}

A simple Evolutionary Algorithm (EA) was developed to evolve designs based on the ranking determined by either the \textit{CLIP-IQA} point-wise evaluation or the \textit{Qwen3-VL-8B-Instruct} pairwise evaluation, with these models determining the fitness score. The EA operated on the parameters of the drawing system (the \textit{genotype}), generating new populations of designs (\textit{phenotypes}) through both pairwise crossover and individual mutation \citep{Mitchell1996,EibenSmith2003}. Tournament selection was used to select parents for recombination, with the tournament proportion size $k \in [0,1]$ available to the artist as an adjustable parameter. The algorithm also allowed the artist to enable elitism, which meant that top ranked individuals from the previous population could compete with the children of the current population. This was controllable through the elitism parameter $e \in [0,1]$ which controlled the proportion of top ranked individuals to be accepted in the current iteration. Parents selected for recombination had their genes subject to random crossover ($p_1, p_2$) to produce a new individual ($c$) using algorithm \ref{alg:1}. 

\begin{algorithm}
\caption{Crossover algorithm used in EA.}
\label{alg:1}
\begin{algorithmic}[1]
\Require Parent solutions $p_1$, $p_2$; mixing parameter $\alpha \in [0,1]$
\Ensure Offspring solution $c$

\For{each variable $x_i$}
    \If{$x_i$ is continuous}
        \State $c_i \gets \alpha p_{1,i} + (1-\alpha) p_{2,i}$
    \ElsIf{$x_i$ is categorical}
        \State Randomly select $c_i$ from $\{p_{1,i}, p_{2,i}\}$
    \EndIf
\EndFor

\State \Return $c$
\end{algorithmic}
\end{algorithm}

The parameter $\alpha \in [0,1]$ was used to weight the arithmetic mean when mixing continuous parameters. The artist was given the option to use a \textit{fixed} $\alpha$, set manually; a \textit{random} $\alpha$; or a \textit{biased} $\alpha$, which biased the mean toward the fitter parent. Once a new population was produced, mutation would be applied to each new individual where each parameter would have a probability $m$ of mutation using algorithm \ref{alg:2}. The artist could control mutation through both the probability of a parameter being mutated $m$ and the extent of mutation for continuous variables $\sigma_m$. The default evolutionary hyperparmeters set for the artist were $k=m=\sigma_m=e=0.1$, $\alpha$ set to biased, elitism active with a population size of 26.

\begin{algorithm}
\caption{Mutation algorithm used in EA.}
\label{alg:2}
\begin{algorithmic}[1]
\Require Individual $c$; mutation probability $m$; Gaussian parameter $\sigma_m$
\Ensure Mutated individual $c'$

\State $c' \gets c$

\For{each parameter $x_i$ in $c'$}
    \State Sample $u \sim \text{Uniform}(0,1)$
    \If{$u < m$}
        \If{$x_i$ is continuous}
            \State $c'_i \gets c'_i + \mathcal{N}(0, \sigma_m^2)$
        \ElsIf{$x_i$ is categorical}
            \State Randomly select $c'_i$ from valid categories
        \EndIf
    \EndIf
\EndFor

\State \Return $c'$
\end{algorithmic}
\end{algorithm}

\subsection{Benchmarking Ranking}

\begin{figure}[!tb]
    \centering
    
    \begin{subfigure}{0.23\textwidth}
        \centering
        \includegraphics[width=\linewidth]{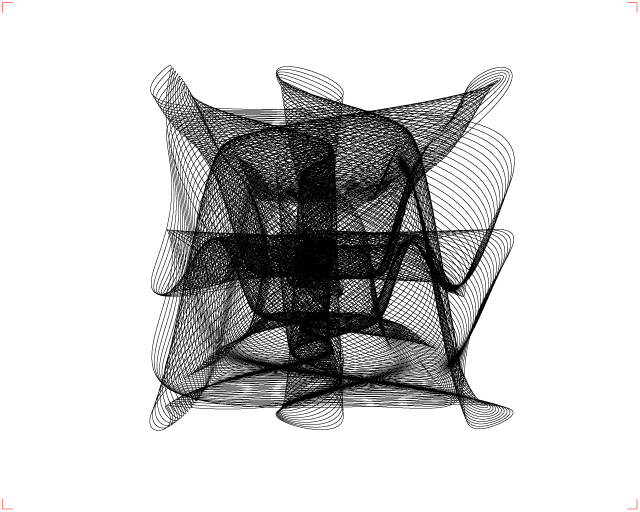}
        \caption{Design 1}
    \end{subfigure}
    \hfill
    \begin{subfigure}{0.23\textwidth}
        \centering
        \includegraphics[width=\linewidth]{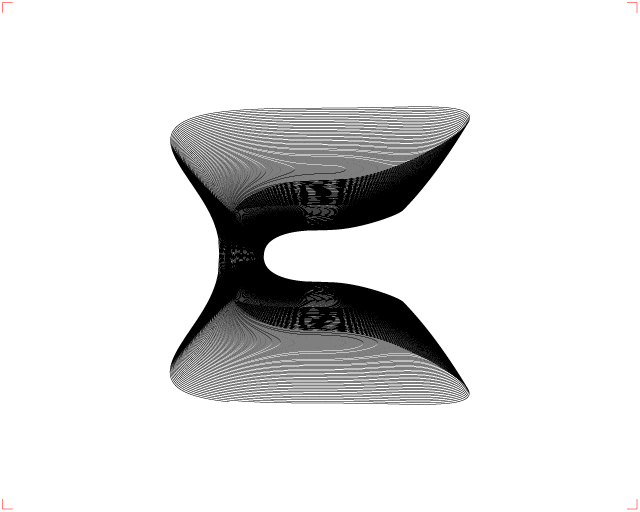}
        \caption{Design 2}
    \end{subfigure}
    \hfill
    \begin{subfigure}{0.23\textwidth}
        \centering
        \includegraphics[width=\linewidth]{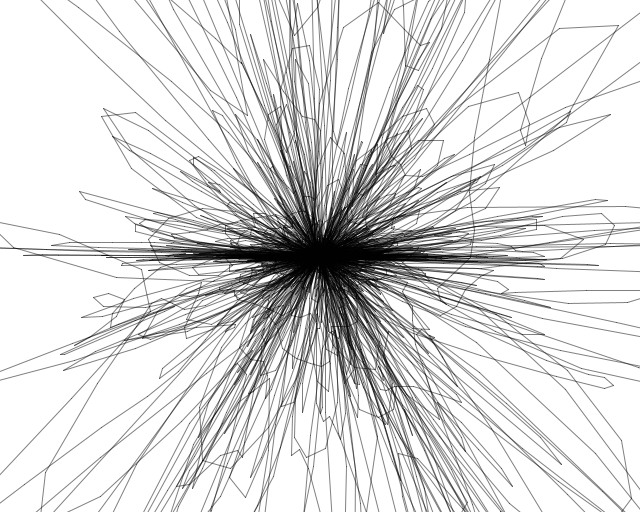}
        \caption{Design 3}
    \end{subfigure}
    \hfill
    \begin{subfigure}{0.23\textwidth}
        \centering
        \includegraphics[width=\linewidth]{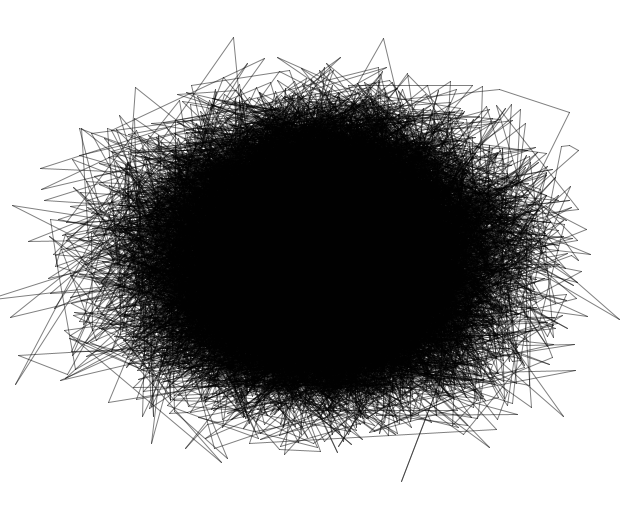}
        \caption{Design 4}
    \end{subfigure}
    
    \caption{Samples from the 100 images used in the benchmark. Designs 1 and 2 are examples of ``quality'' designs, defined by their structure and aesthetic. Designs 3 and 4 are examples of ``low-quality'' designs, defined by their messy appearance and lack of structure.}
    \label{fig:benchmark_examples}
\end{figure}

To quantitatively measure the performance of both methods in ranking the generated designs, a custom benchmark was built to compare each technique's ranking to that of the artist's. While this does not provide an exhaustive evaluation of each technique's performance in ranking the aesthetics of non-representational images, it does provide insight into their ability to align to an artist's aesthetic style through custom prompts. To construct this ranking, a web interface was built that enabled the artist to compare pairs of images --- selecting either a winner or a draw --- while the \textit{Glicko} system was used to estimate the ranking of 100 designs. The ranked designs included a mix of randomly generated and curated examples, as purely random generation often produced a large proportion of low-quality designs. In total, the artist performed 1500 comparisons to build the ranking. An example of these designs can be seen in Figure~\ref{fig:benchmark_examples}. 

To compare the computed rankings to the artist's rankings, the following metrics were used: 

\begin{enumerate}
    \item Spearman's rank-order correlation (SPCC) \citep{spearman1987proof}, a nonparametric measure of rank correlation that assesses the strength and direction of the association between two ranked variables used in similar studies \citep{xiong2024image,wang2023exploring}. The score is measured between $[-1,1]$, where 1 indicates perfect positive rank correlation, -1 indicates perfect negative rank correlation, and 0 indicates no monotonic association between the rankings.
    \item Kendall’s tau \citep{kendall1938new}, a nonparametric measure of rank correlation that quantifies the strength and direction of association between two rankings by comparing the number of concordant and discordant pairs of items. The score is measured between $[-1,1]$, where 1 indicates perfect agreement between the rankings, -1 indicates perfect disagreement, and 0 indicates no association between the rankings.
    \item The Jaccard top-k measure quantifies the similarity between two rankings by computing the Jaccard index (size of the intersection divided by the size of the union) of their sets of top-k items. The score ranges from $[0,1]$ where 0 indicates no overlap and 1 complete overlap. For these experiments $k = [5, 10, 20]$ to test agreement of top ranked designs.
\end{enumerate}

The following experiments were run: (1) Using the structural complexity, MC complexity, fractal complexity and fractal dimension scores developed in previous work to measure aesthetics \citep{mccormack2022complexity}, (2) CLIP-IQA using the prompt pairing of \textit{Good Design} and \textit{Bad Design}, which through trial-and-error produced the best visual results, (3) Pairwise comparison with \textit{Qwen3-VL-8B-Instruct}, both with a reasoning prompt (see Figure~1) and without a reasoning prompt and (4) Pairwise comparison with \textit{Qwen3-VL-8B-Thinking}. The prompt used in these benchmarks was designed by the artist and can be seen in Figure 5.

\begin{figure}
\begin{tcolorbox}[
    title=Artist Prompt,
    colback=gray!5,
    colframe=gray!60,
    boxrule=0.5pt,
    arc=2pt
]

You are a highly experienced abstract art critic.
You will be given two images of monochrome line drawings, you must apply your artistic judgement to decide which image is more artistic.

Task:
Critique each image from the perspective of an experienced art critic. If you cannot make a decision return a draw value of `3'.
Output a one sentence description of each image, a single sentence regarding your reasoning and finally a single digit corresponding to which image is better ONLY `1' or `2' or `3' for draw.

\end{tcolorbox}
\label{fig:artist-prompt}
\caption{Prompt used by the artist.}
\end{figure}

\begin{table}[ht]
    \centering
    \setlength{\tabcolsep}{3pt}
    \begin{tabular}{@{}p{4cm}ccccc@{}}
    \toprule
     Method & SPCC   &  Tau & J@5 & J@10 & J@20 \\
     \midrule
     Structural Complexity & 0.36 & 0.22 & 0.00 & 0.00 & 0.05 \\
     MC Complexity & 0.43 & 0.27 & 0.00 & 0.00 & 0.05 \\
     Fractal Complexity & 0.07 & 0.08 & 0.00 & 0.00 & 0.03 \\
     Fractal Dimensions & 0.06 & 0.09 & 0.00 & 0.00 & 0.03 \\
     \noalign{\vskip 4pt}
     \noalign{\vskip 4pt}
     Point-wise CLIP-IQA & 0.79 & 0.58 & 0.00 & 0.05 & 0.29 \\
     \noalign{\vskip 4pt}
     \noalign{\vskip 4pt}
     Pairwise VLM Instruct w/out Reasoning Prompt & 0.64 & 0.45 & 0.00 & 0.00 & 0.03 \\
     Pairwise VLM Instruct@10 & 0.75 & 0.55 & 0.02 & 0.1 & 0.24 \\
     Pairwise VLM Instruct@20 & 0.78 & 0.58 & 0.01 & 0.08 & 0.25 \\
     Pairwise VLM Instruct@50 & 0.79 & 0.59 & 0.00 & 0.07 & 0.27 \\
     Pairwise VLM Instruct & \textbf{0.8} & \textbf{0.6} & 0.00 & 0.05 & 0.29 \\
     \noalign{\vskip 4pt}
     \noalign{\vskip 4pt}
     Pairwise VLM Thinking & 0.76 & 0.55 & 0.00 & 0.11 & 0.33 \\
     \bottomrule
    \end{tabular}
    \caption{Scores from benchmarks. Including four measures from previous work, the CLIP-IQA method, pairwise VLM without a reasoning prompt, pairwise instruct VLM at different level of comparisons: 10\%, 20\%, 50\% and complete and thinking VLM with complete comparisons.}
    \label{tab:benchmark}
\end{table}

The scores from the benchmarks can be seen in Table \ref{tab:benchmark}. From these results, it's clear that all DL methods better align to the artist's ranking compared to traditional proxies \citep{mccormack2022complexity}, with scores above 0.75 suggesting high correlation. Amongst the DL approaches, the \textit{pairwise VLM instruct} method scores the highest but with no significant difference to the CLIP-IQA method. This is particularly apparent when considering the top-k performance, where all approaches fail to strongly capture the top preferences of the artist. Regarding the effect of the number of comparisons $n$ on Glicko pairwise ranking performance, the table shows that sampling only 50\% of comparisons results in just a 1\% drop in performance while reducing compute cost by 50\%. However, this relationship begins to break down at around 10\% of the total comparisons.

\subsection{Time Benchmarking}

The above experiments were run on a Nvidia RTX5090 GPU with \textit{Flash Attention} and inference batching. All experiments, excluding the pairwise models, took less than a second to compute the rankings for 100 designs. For pairwise calculations, it took approximately 16 minutes using \textit{Qwen3-VL-8B-Instruct} and 45mins with \textit{Qwen3-VL-8B-Thinking}, to compute all 4950 pairwise combinations of the 100 designs. As there is no significant difference between the scores of the pairwise VLM and point-wise CLIP-IQA methods, these benchmarks provide little motivation to adopt the current implementation of VLM-based pairwise aesthetic evaluation, given the substantial increase in compute time, even with pairwise sampling.  

\section{Discussion}

\subsection{Artist's Experience}

Having discussed the quantitative aspects of the study, we now turn to the artist's perceptions on using the system as a way to help discover new and aesthetically suitable designs. The artist worked with a web-based interface, able to select between the different fitness evaluation methods and the parameters discussed.

The artist commented that they did not feel changing the text prompts to CLIP-IQA gave a strong sense that they were being followed by the system. For example, prompts such as ``insect-like form'' and ``not insect-like form'' did not seem to evolve forms that resembled insects at all, offering few discernible differences over the defaults of ``good design'' and ``bad design''. This made CLIP-IQA more conceptually difficult to control, or at least for the artist to understand how to change what the system was \textit{evolving for} based on the antonym prompt pairs.

Using the VLM however, the artist felt more control over the designs produced and commented that VLM was indeed able to find a number of designs that could potentially be useful artistically. There was a stronger \textit{perceived} connection between the text prompt used and the evolutionary direction, despite the VLM's tendency to often evolve towards designs that were too dark and dense after several generations. Even being specific about avoiding designs that were too dense in the instructional prompt did not fully elevate this problem. Another discovery was that trying to be too specific in the prompting did not result in better designs or the system's ability to evolve towards specific forms listed in the prompt. For example, the ``art critic'' prompt (Figure 5) generated better results than prompts referring to specific figurative forms, such as insect, bird or tree-like -- all forms the system can (more-or-less) generate rough approximations of.

Lastly, the VLM is a general system not trained on artist's preferences or any of the designs from the system evaluated. The artist felt that if the system was able to better understand what specific terms meant in relation to the design system, they would better be able to articulate in language the general evolutionary direction required.

\begin{figure}[!tb]
    \centering
    
    \begin{subfigure}{0.23\textwidth}
        \centering
        \includegraphics[width=\linewidth]{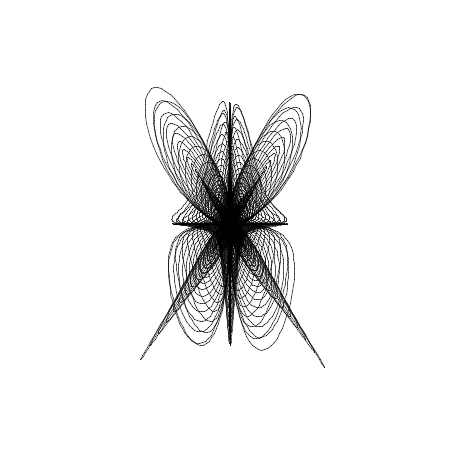}
        \caption{Butterfly 1}
    \end{subfigure}
    \hfill
    \begin{subfigure}{0.23\textwidth}
        \centering
        \includegraphics[width=\linewidth]{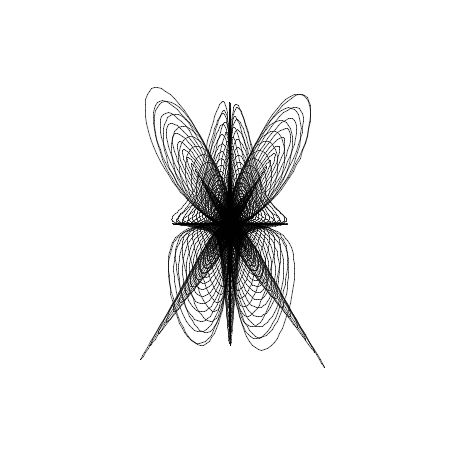}
        \caption{Butterfly 2}
    \end{subfigure}
    \hfill
    \begin{subfigure}{0.23\textwidth}
        \centering
        \includegraphics[width=\linewidth]{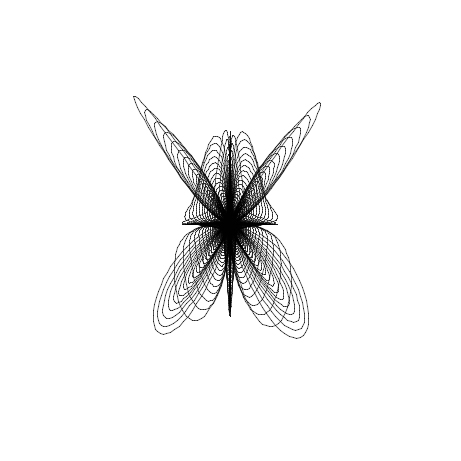}
        \caption{Butterfly 3}
    \end{subfigure}
    \hfill
    \begin{subfigure}{0.23\textwidth}
        \centering
        \includegraphics[width=\linewidth]{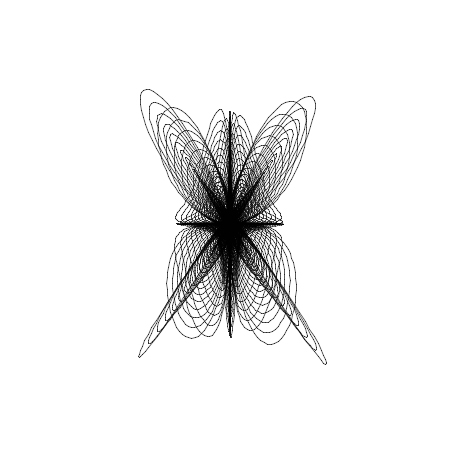}
        \caption{Butterfly 4}
    \end{subfigure}
    
    \caption{Generations discovered by the pairwise VLM evolving to designs that resemble a butterfly.}
    \label{fig:butterflys}
\end{figure}

\subsection{Point-wise vs Pairwise}

This work presents an early investigation of two IAQA approaches for non-representational images: a point-wise method based on CLIP-IQA and a pairwise method using the \textit{Qwen3-VL-8B-Instruct} model. While prior research \citep{yannakakis2018ordinal} suggests that pairwise comparisons may better align with human aesthetic judgement --- given its inherently ordinal nature --- our quantitative results indicate that neither approach consistently outperformed the other, with both techniques achieving a high correlation to the artist's preferences. This outcome may stem from several factors, including the relatively small scale of the VLM underlying the pairwise evaluations \citep{kaplan2020scaling}, or limitations of the Glicko ranking procedure itself. As such, further investigation is required before drawing firm conclusions about the comparative performance of the two approaches. That said, the current quantitative results favour the point-wise method, primarily due to its substantially lower computational cost.

\begin{figure}[!tb]
    \centering
    
    \begin{subfigure}{0.23\textwidth}
        \centering
        \includegraphics[width=\linewidth]{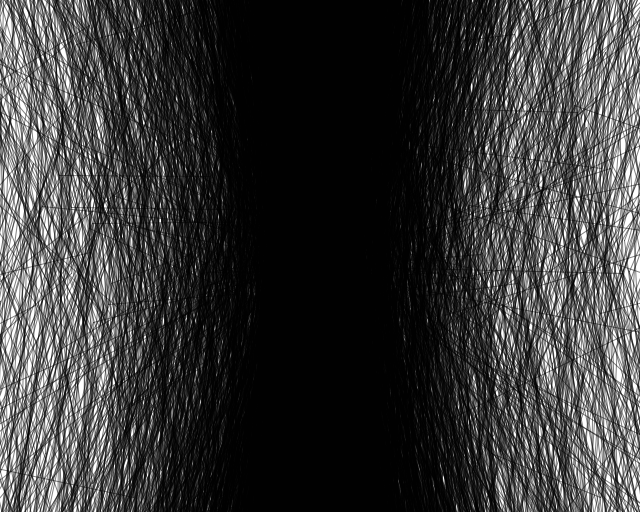}
        \caption{CLIP-IQA Score: 0.86}
    \end{subfigure}
    \hfill
    \begin{subfigure}{0.23\textwidth}
        \centering
        \includegraphics[width=\linewidth]{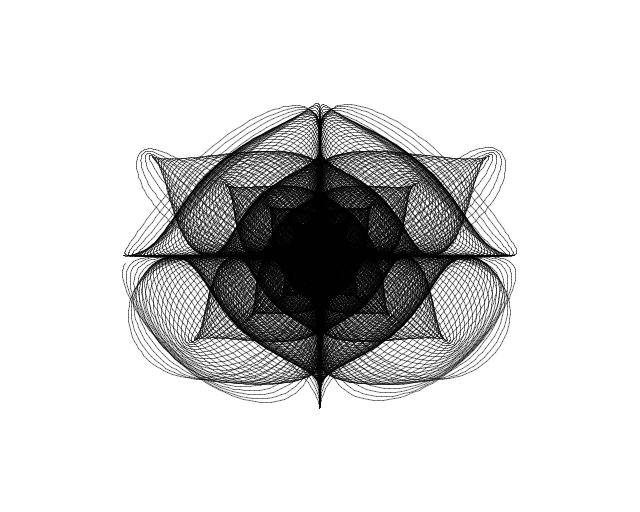}
        \caption{CLIP-IQA Score: 0.84}
    \end{subfigure}
    
    \caption{Rankings from the CLIP-IQA method demonstrating how one `bad' design is scored higher than a `good' design, with little options for the artist to prevent this.}
    \label{fig:CLIP-IQA}
\end{figure}

Conversely, from the artist’s perspective, the pairwise approach was preferred, as it afforded a stronger sense of control over the evaluation process. This preference is unsurprising, given that some degree of user control is frequently emphasised as central to AI creativity support tools \citep{santo2023focusing,krol2025}. The flexibility in prompt design enabled by the VLM allowed the artist to articulate their vision more precisely. This is illustrated in Figure~\ref{fig:butterflys}, which shows phenotypes evolved from a prompt instructing the model to favour butterfly-like designs. Furthermore, CLIP-IQA sometimes ranked weaker images above stronger ones (Figure~\ref{fig:CLIP-IQA}), allowing poorer designs to carry over into the next generation. While the pairwise method also made occasional errors, comparing each design against many others reduced the impact of individual misjudgements, since rankings reflected overall performance rather than a single score. 

Finally, this work suggests that both methods would benefit from greater personalisation. The top-k benchmarks show that neither approach reliably captured the artist’s highest-ranked preferences, and the artist also remarked that the models felt overly general rather than tailored to their aesthetic sensibilities. Although recent studies have begun to explore fine-tuning models for aesthetic evaluation \citep{xiong2024image}, our results reinforce the need for approaches that support personalisation --- particularly in small-data scenarios \citep{abuzuraiq2024towards} --- so that artists can meaningfully adapt these systems to their own practice.

\section{Limitations \& Future Work}

While this work begins to explore DL approaches to evaluating non-representational images for EC, it does so in the context of an artist-centred study. While this offers practical insight into the performance of both techniques, further work is needed to develop stronger benchmarks and conduct broader evaluations in order to obtain more reliable and generalisable assessments of their performance .

Future work will also aim to experiment with different VLMs to compare how changes in models affect performance, specifically model complexity. Additionally, different ranking algorithms such as the Bradley–Terry model \citep{bradley1952rank}. 

Finally, future work should also investigate the value of VLMs explaining their aesthetic choices \citep{llano2022explainable} and whether this adds any value to the interaction.

\section{Conclusion}
In this paper we have explored the novel use of pre-trained IAQA models deployed as automated fitness measures in creative evolutionary applications. We tested models with a professional artist, using a bespoke generative drawing system they had developed as part of their established creative practice. Our results showed that both CLIP-IQA and VLM models performed well at following the artist ranking of designs, with the pairwise VLM Instruct model narrowly coming out on top. Both models performed better than more simple measures of complexity.

While both CLIP-IQA and the VLM were roughly similar in performance, the artist much preferred the VLM and its richer possibilities for customisation of the prompt in determining the evolutionary direction. This preference comes at a significant computational (and hence time) cost over CLIP-IQA however.

In a landscape increasingly flooded with ``AI slop'' \citep{mahdawiAIgeneratedSlopSlowly2025} and general anxiety from the creative community regarding generative AI \citep{taitArtistsAIDilemma2024,bakareArtThatCan2024}, artist-designed creative systems offer a valuable and important alternative. Bespoke artistic systems are still capable of creating results not achievable with commercial prompt-to-image models. Additionally, they express the personal over the statistical average. We hope our study has illustrated ways in which deep-learning and contemporary AI models can still play a valuable role in supporting the creative possibilities for individual artists.

\section{Acknowledgements}

The research was supported by an Australian Research Council Discovery Project Grant DP250100230.

\bibliographystyle{iccc}
\bibliography{iccc}

\end{document}